\documentclass{article}

\usepackage{microtype}
\usepackage{graphicx}
\usepackage{subfigure}
\usepackage{booktabs} % for professional tables
\usepackage{xcolor}
\usepackage{graphicx}
\usepackage{multirow,times,epsfig,pdflscape,graphicx,amsmath,amssymb}
\usepackage{float}
\usepackage{amsmath,amssymb,amsfonts}
\usepackage{graphicx,textcomp}
\usepackage{mathrsfs}
\usepackage{amsmath}
\usepackage{empheq}
\usepackage{amssymb}
\usepackage{rotating}

\usepackage[utf8]{inputenc} % allow utf-8 input
\usepackage[T1]{fontenc}    % use 8-bit T1 fonts
\usepackage{hyperref}       % hyperlinks
\usepackage{url}            % simple URL typesetting
\usepackage{booktabs}       % professional-quality tables
\usepackage{amsfonts}       % blackboard math symbols
\usepackage{nicefrac}       % compact symbols for 1/2, etc.
\usepackage{microtype}      % microtypography
\usepackage{lipsum}
\usepackage[noadjust]{cite}
\usepackage{amsmath,amssymb,amsfonts}

\usepackage{array, caption, tabularx,  ragged2e,  booktabs}
\usepackage{longtable}
\usepackage{textcomp}
\usepackage{siunitx}
\usepackage{array}
\usepackage[linesnumbered,ruled,vlined]{algorithm2e}
\usepackage{romannum}
\usepackage{graphicx}
\usepackage{amsthm}
\usepackage{graphicx}
\usepackage{xcolor}

\usepackage{times}
\usepackage{epsfig}
\usepackage{amsmath}
\usepackage{amssymb}
\usepackage{appendix}
\usepackage{etoolbox}
\usepackage{soul}
\usepackage{tabularx}
\usepackage{caption}

\usepackage[capitalize]{cleveref}

\usepackage[inline]{enumitem}

\usepackage{hyperref}

\BeforeBeginEnvironment{appendices}{\clearpage}

\newcolumntype{Y}{>{\centering\arraybackslash}X}
\def\lce{{\mathcal{L}_{\rm CE}}}

\DeclareMathOperator*{\minimize}{minimize}

\definecolor{todocolor}{RGB}{200,120,120}
\sethlcolor{todocolor}

\usepackage{graphicx} 
\usepackage{caption}

\def\D{{\mathcal{D}}}

\def\ood{{OOD}}
\def\oods{{OOD~}}

\newcolumntype{M}[1]{>{\centering\arraybackslash}m{#1}}
\newcolumntype{N}{@{}m{0pt}@{}@{}}

\graphicspath{{images/}}
\graphicspath{{./imgs/}}

\usepackage{array, caption, multirow, tabularx,  ragged2e,  booktabs}
\usepackage[accepted]{icml2021}

\usepackage{hyperref}

\usepackage{icml2021}

\icmltitlerunning{ICML 2021 Workshop on Uncertainty and Robustness in Deep Learning}

\begin{document}

\twocolumn[
\icmltitle{PnPOOD
: Out-Of-Distribution Detection for Text Classification via Plug and Play Data Augmentation}

%using data augmentation from Plug and Play models in Natural language processing\ramya{needs refinement}}

% It is OKAY to include author information, even for blind
% submissions: the style file will automatically remove it for you
% unless you've provided the [accepted] option to the icml2021
% package.

% List of affiliations: The first argument should be a (short)
% identifier you will use later to specify author affiliations
% Academic affiliations should list Department, University, City, Region, Country
% Industry affiliations should list Company, City, Region, Country

% You can specify symbols, otherwise they are numbered in order.
% Ideally, you should not use this facility. Affiliations will be numbered
% in order of appearance and this is the preferred way.

% You may provide any keywords that you
% find helpful for describing your paper; these are used to populate
% the "keywords" metadata in the PDF but will not be shown in the document

% \vskip 0.3in
% ]

% this must go after the closing bracket ] following \twocolumn[ ...

% This command actually creates the footnote in the first column
% listing the affiliations and the copyright notice.
% The command takes one argument, which is text to display at the start of the footnote.
% The \icmlEqualContribution command is standard text for equal contribution.
% Remove it (just {}) if you do not need this facility.

%\printAffiliationsAndNotice{}  % leave blank if no need to mention equal contribution
% standard text.

\icmlsetsymbol{equal}{*}
\begin{icmlauthorlist}
\icmlauthor{Mrinal Rawat}{tcs}
\icmlauthor{Ramya Hebbalaguppe}{tcs}
\icmlauthor{Lovekesh Vig}{tcs}
\end{icmlauthorlist}

\icmlaffiliation{tcs}{TCS Research \& Innovation}

\icmlcorrespondingauthor{Mrinal Rawat}{rawat.mrinal@tcs.com}
\icmlcorrespondingauthor{Ramya Hebbalaguppe}{ramya.hebbalaguppe@tcs.com}

% You may provide any keywords that you
% find helpful for describing your paper; these are used to populate
% the "keywords" metadata in the PDF but will not be shown in the document
\icmlkeywords{OOD detection, Chatbots, PPLM}

\vskip 0.3in
]

\printAffiliationsAndNotice{\icmlEqualContribution} % otherwise use the 

\begin{abstract}

 While Out-of-distribution (OOD) detection has been well explored in computer vision, there have been relatively few prior attempts in OOD detection for NLP classification.  In this paper we argue that these prior attempts do not fully address the OOD problem and may suffer from data leakage and poor calibration of the resulting models. We present PnPOOD, a data augmentation technique to perform OOD detection via  out-of-domain sample generation using the recently proposed Plug and Play Language Model\cite{DBLP:journals/corr/abs-1912-02164}. Our method generates high quality discriminative samples close to the class boundaries, resulting in accurate OOD detection at test time. We demonstrate that our model outperforms prior models on OOD sample detection, and exhibits lower calibration error on the 20 newsgroup text and Stanford Sentiment Treebank dataset \cite{Lang95, socher-etal-2013-recursive}. We further highlight an important data leakage issue with datasets used in prior attempts at OOD detection, and share results on a new dataset for OOD detection that does not suffer from the same problem.

% \ramya{Lovekesh, we need to shorten this as its way too long}
\end{abstract}

%\footnote{Trivia: PnPOOD is an acronym for Plug and Play data-augmentation based OOD detection. The name PnPOOD is a homonym for Pea-pod that separates the pea (In-Distribution data) from the open world (OOD samples). Since we found the pod to be a good analogy to class boundary, we name the algorithm, PnPOOD}

\section{Introduction}
\label{sec:intro}
Most conversational agents deployed for enterprise applications have a specific purpose, such as assisting employees to answer questions about HR policies or resolving IT infrastructure issues. These agents are trained to answer queries with fixed intents from a particular domain via a text classifier that classifies user queries into one of several pre-defined intents. However, deep NLP models employed for intent classification in conversational systems are susceptible to improper responses to user queries due to overconfident predictions on out-of-domain (OD) test samples ~\cite{hendrycks2016baseline, 10.5555/3305381.3305518}. Automatic detection and redirection of OOD samples to other bots or for manual intervention would improve user experience  and  enhance trust in such systems.  Prior attempts at OOD detection for text/image classification have varied from entropy maximization based methods ~\cite{lee2017training,10.5555/3327757.3327819}, data augmentation techniques~\cite{hendrycks2018benchmarking, patel2021manifold,hein2019relu} and uncertainty quantification methods\cite{10.5555/3295222.3295387, gal2016dropout}. However, often these methods ignore the impact of OOD detection on the calibration error of the intent classifier. A further problem with OOD detection datasets reported in prior work is that the OOD samples generated for detection training and those encountered during testing may belong to  overlapping domains. To this end, we present PnPOOD, an OOD detection algorithm that significantly outperforms prior methods, both on OOD detection metrics and model calibration error on text classification.  The technique utilizes a PPLM model ~\cite{DBLP:journals/corr/abs-1912-02164}, a computationally light technique to guide sentence generation for OOD samples, which are then used to train the OOD classifier. To ensure high quality OOD sample generation, we provide initial guiding tokens from in-domain (ID) samples. We demonstrate the superiority of our technique on the SST dataset~\cite{socher-etal-2013-recursive}, on which all prior results have been reported in the recent past\cite{hendrycks2016baseline, hendrycks2018benchmarking}. Further, we discovered a data leakage issue with the previous approaches that has so far been overlooked. Therefore, in order to provide an unbiased assessment of our model, we also conducted experiments on a dataset carved out of the $20$ newsgroup\cite{Lang95} dataset ensuring no data leakage between train and test OOD samples.   
%\todo{Ramya to add all references of organisers}

The primary contributions of this work are: (a) We propose PnPOOD, a new technique for generating and training OOD detectors on OOD samples that are close to the class decision boundaries. (b) Results show our technique outperforms all existing methods for OOD detection on text classification in the recent past, both in terms of detection accuracy and model calibration

%The OOD samples should be forced to a uniform distribution achieved by reducing the KL divergence between a uniform distribution and OOD samples or explicitly highering the entropy on OOD samples.

%\ramya{ Lovekesh to write about parent child chat bot and also computational efficiency of the our approach. We are not using OOD data.}

%We find it very difficult to provide a complete list of
%references although a sincere attempt has been made to quote the most relevant works for a short paper submission. 

\section{Related Work}
\label{sec:rel_work}
The recent techniques fine-tune hyper-parameters on a validation set to optimize OOD detection, for example, Hendrycks and Gimpel (MSP) \cite{hendrycks2016baseline} use the maximum confidence scores from a Softmax output as a detection score, which in turn is used to classify \oods samples. ODIN utilizes temperature scaling with input perturbations using the \oods validation dataset to tune hyper-parameters~\cite{liang2017enhancing}. However, hyperparameters tuned with one \ood dataset is found not to generalize to other datasets. The sorted Euclidean distance between the input and the k-nearest training samples has also been used as a detection score~\cite{zhang2006svm}. The likelihood ratio method for deep generative model corrects for confounding background statistics and is used as an effective OOD detection method in image classifiers. A background model is trained using perturbed IND samples employing a Liklihood ratio which enhances OOD detection performance ~\cite{NEURIPS2019_1e795968}. Lee et al. \cite{10.5555/3327757.3327819} propose detecting \oods samples by training a logistic regression detector on the Mahalanobis distance vectors calculated between test images' feature representations and the class conditional Gaussian distribution at each layer. ~\cite{lee2017training} generate synthetic images sampled from the low density boundary regions of the in-distribution space. For generating synthetic OOD samples around uniform distribution, they propose using GANs. OOD samples can be forced to have an uniform distribution by minimizing the Kullback-Leibler divergence between the model generated probabilities on OOD samples and a uniform distribution. The OOD detectors listed before are demonstrated mainly on visual tasks, in particular, on image classification. Our focus has been to advance the state-of-the-art in OOD detection in text classification where the research efforts are scant and our technique is focussed to intent classification in natural language processing using Natural language generation that we describe in future.

%in the training objective that maximize entropy for  \oods samples

\begin{figure}[!ht]
\centering
\includegraphics[width=1.0\linewidth]{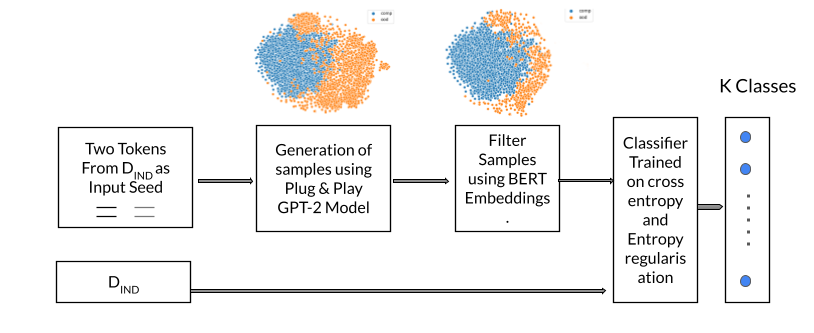} 
\caption{An OOD classifier is trained on OOD samples generated using Plug and Play Language model and IND data. At a high level we employ entropy regularisation on OOD samples. Intuitively, when we force the samples to have highest entropy, OOD samples are closer to the uniform distribution (zero confidence). A sample is considered OOD based on observing the probability vector at the classifiers output being low on all the elements of a prediction vector. Also, our technique filters the subset of OOD samples generated from PPLM to concentrate around the class boundary thereby improving the OOD performance. }
\label{fig:architecture}
\end{figure}

\section{Methodology}
\label{sec:methodology}

%\todo{write a simple english description}
%We have a dataset, $\D_\text{ind}^\text{train} = (x_n, y_n)^N_{n=1}$, comprising of $N$ training samples where $\D_\text{ind}^\text{train}$. Here, $X\in\X$ is defined on input space, and $Y\in\Y = \{1,\ldots,K\}$ corresponding labels. A classifier $f_{\theta}:\mathcal{X} \rightarrow \mathcal{Y}$ is trained on the IND samples drawn from the marginal distribution $\cP_X$ of $X$ derived from the joint distribution $\cP_{X,Y}$, $\theta$ are the parameters of DNN. While testing, the inputs are drawn from a conditional mixture distribution $\cM_{X | Z}$ where $Z \in \{0,1\}$, such that $\cM_{X \mid Z=0} = \cP_X$, and $\cM_{X \mid Z=1} = \cQ_X$. $\cQ_X \nsim \cP_X$ as \oods distributions, and $Z$ is a latent variable to encode IND if $Z=0$ and OOD if $Z=1$ . %In a multi-class setting, a sample can be treated as OOD if the output probability vector from $f_\theta$ for a given input is low for all elements of $\mathcal{Y}$ indicating low/zero confidence in case of uniform distribution.

 For formulation/definition of OOD detection in a supervised classification setting, See \cite{liang2017enhancing}. We propose a practical approach to post-hoc OOD detection i.e. situations where OOD detection has to be incorporated into an existing classification model without model retraining. The dataset $D$ comprises of sentences from  $N$ different domains. Out of these $N$ domains, $k$ domains are treated as IN-DISTRIBUTION (IND) while the rest of the $N-k$ domains are treated as OUT-DISTRIBUTION (OOD). Given a set of IND sentences $D_{ind}$   =  ${(x_1, y_1), (x_2, y_2), \ldots  (x_n, y_n)}$, we train a network using the following loss function:

\begin{align}
	\mathcal{L} = \minimize_{\theta} 
	\hspace{0.35cm} 
	\mathbb{E}_{(x, y) \sim  D_\text{ind}^\text{train}} [\lce( y_{in},f_{\theta}(x))] \nonumber\\
	+ \alpha \cdot \mathbb{E}_{(x_{ood}^{PPLM}) \sim  D_\text{out}^\text{OOD}} [\mathcal{L_{E}}(f_{\theta}(x_{ood}^{PPLM}), U(y))]
	%\end{align}
	\label{equ:loss}
\end{align}

where $\lce$ is the cross entropy loss, $f_{\theta}(x)$ denotes the softmax output prediction of neural network for an input sample $x$. We obtain least loss when $\alpha = 1$ is used our experiments. $x_{ood}^{PPLM}$ is the OOD sample generated from our data augmentation using PPLM light weight models as detailed in Fig \ref{fig:architecture}. $D_\text{out}^\text{OOD}$ is the OOD dataset.  $\mathcal{L_{E}}$ is a regularisation that tries to minimize the loss between output probability vector  of an OOD sample $x_{ood}^{PPLM}$ to an uniform distribution inspired from \cite{lee2017training}. Note that it is harder to model a complete test distribution as the OOD space is infinite. However, we can use heuristics for detecting test distribution using just the representations from only IND data. Our approach is one such that considers taking the low density samples at the class boundaries of IND distribution to model OOD. Section \ref{subsec:dataAugOOD} throws light on OOD sample generation to obtain $D_\text{out}^\text{OOD}$ for training an OOD detector as explained in Equation \ref{equ:loss}. We notice, that the training procedure using our OOD samples significantly helps in boosting the OOD detection performance.

%\subsection{Entropy Regularization}
%Write about entropy reg.. and eqns....

\subsection{PPLM based OOD data augmentation}
\label{subsec:dataAugOOD}

We propose a novel technique of generating the sentences which are close to the IN-Domain samples. We hypothesize that the samples which are closer in the embedding space to the IND cluster boundary are more relevant for discriminating between IND and OOD samples. This is particularly important as the OOD space is huge and there exists no principled way of generating samples that are representative of the entire OOD space. These generated sentences form proxy for OOD which is used for entropy regularisation, thus improving the OOD detection performance. Fig \ref{fig:architecture} depicts the OOD sample generation and training procedure.

While large scale language models like GPT-2~\cite{radford2019language} have shown remarkable performances in Natural Language generation (NLG), guiding the generation for a task often required compute intensive fine tuning of the full model. The recent proposed Plug and Play Language Models (PPLM)~\cite{DBLP:journals/corr/abs-1912-02164}  utilize an attribute model in order to guide generation without fine tuning the large model. In this work, we utilize PPLM to generate new samples whose BERT\cite{DBLP:journals/corr/abs-1810-04805} embeddings are closer to IN-DOMAIN cluster boundary. To generate the samples using the PPLM, we provide the initial input seed and the bag-of-words which is used to control the generation. We take the initial two tokens from random samples in $D_{ind}$ as input seed and generate sentences directed towards $D_{ood}$ using the light weight PPLM training procedure. The bag-of-words are extracted from the out-domain dataset since they guide the model to generate the sentences which are out-of-distribution. For example,

\textbf{Original Sentence:} ``\underline{This article} includes answers what options have for software intel based unix system''

%\ramya{this sentence seems incorrect gramatically, Mrinal please check - after options, is it "you have"}

\textbf{Input:} \textcolor{red}{[Science]} $<$This article$>$

\textbf{Output:} \underline{This article} explores a recent study on a large scale of global \textcolor{orange}{climate} system climate change, which finds no direct evidence of the \textcolor{orange}{Earth's} climate warming. 

Words from $D_{ood}$  (from science domain):

\texttt{astronomy, atom, biology, cell, chemical, chemistry, earth, climate}

Despite controlled generation, we observed that some generated samples were quite far from the IND samples. To overcome this problem, we filter the sentences based on the BERT embeddings \cite{wolf-etal-2020-transformers}. We first find the IND cluster center $C_{ind}$. We then measure the distance of each sentence embedding with the cluster center and consider only the ones which are closer to the cluster boundary. Refer Alg. \ref{algo_PeaPOOD} in Supplemental material for pseudocode of our proposed OOD sample generation including OOD sample filtering. 

% The threshold is $>$ distance of center to boundary ($d$) and $<$ d+10, where we obtained threshold to be $10$ based on fine-tuning.
% The sentences which are far away from the $D_{ind}$  are rejected. \ramya{This sentence needs to be changed to an equation and clearly define variables used}

% use the equation of an ellipse

\subsection{Post-hoc calibration}
The modern DNNs produce overconfident decisions, to overcome this, in addition to OOD detection, we apply Dirichlet calibration\cite{kull2019beyond} as a post-hoc calibration technique. They assume $\hat{p}(X | Y=j) \sim $ Dirichlet$(\alpha^{(j)})$, where $\alpha^{(j)} \in \mathbb{R}^K$. They propose a new regularisation method called Off-Diagonal regularisation, given by $ODIR = \frac{1}{k*(k-1)}\sum_{i \neq j}w_{i,j}^2$. This post-hoc calibration improves the consonance of predicted probabilities with the accuracies produced by the classifier. Dirichlet calibration can be thought of log-transforming the uncalibrated probabilities, followed by one linear layer and softmax; this simple procedure is known to outperform temperature/vector scaling to produce well-calibrated scores~\cite{hinton2015distilling}.

% We experiment using the following datasets:

% \textbf{P1}: 

% $D_{ind}$ = $<$Computer$>$

% $D_{ood}$ = $<$Sports, Politics$>$

% $D_{ood - ER}$ = $<$Science, Religion$>$ # Sentences drawn from the original dataset

% $D_{ood-ER-PPLM}$ = $<$Science, Religion$>$ # Sentenced generated using PPLM

% \textbf{P2}: 

% $D_{ind}$ = $<$Sports$>$

% $D_{ood}$ = $<$Computer, Politics$>$

% $D_{ood-ER}$ = $<$Science, Religion$>$ # Sentences drawn from the original dataset

% $D_{ood-ER-PPLM}$ = $<$Science, Religion$>$ # Sentenced generated using PPLM

% \textbf{P3}: 

% $D_{ind}$ = $<$Politics$>$

% $D_{ood}$ = $<$Computer, Sports$>$

% $D_{ood-ER}$ = $<$Science, Religion$>$ # Sentences drawn from the original dataset

% $D_{ood-ER-PPLM}$ = $<$Science, Religion$>$ # Sentenced generated using PPLM

% \textbf{Models Used:}

% \begin{itemize}
%     \item \textbf{MSP:} Trained on $D_{ind}$ and tested on $D_{ood}$
    
%     \item \textbf{Entropy Reg:} Trained on $D_{ind}$ and $D_{ood-ER}$ , and tested on $D_{ood}$

%     \item \textbf{Entropy Reg + PPLM:} Trained on $D_{ind}$ and $D_{ood-ER-PPLM}$ , and tested on $D_{ood}$
% \end{itemize}

% Out of these we take 3 domains as 
%\end{abstract}

\section{Experiments}
\label{sec:experiments}

 \textbf{SST} We evaluated our approach by training with SST \cite{socher-etal-2013-recursive}  dataset, and tested on Multi30K\cite{W16-3210} and SNLI\cite{snli:emnlp2015} OOD datasets. The Stanford Sentiment Treebank (SST) dataset consists of movie reviews
expressing positive or negative sentiment. SNLI is a dataset of predicates and hypotheses for natural language inference and Multi30K is a dataset of English-German image descriptions.

\textbf{20 Newsgroups} We evaluate our approach and compared against  other baselines on The 20 Newsgroups \cite{Lang95} dataset. This dataset consists of approximately 20000 newsgroups documents divided into 20 groups. These 20 newsgroups correspond to different topics but some of the overlapping topics can be merged yielding 6 major domains. Table ~\ref{stats} illustrates domain and group information present in Appendix \ref{appendix:eval}.

\setlength{\tabcolsep}{2pt}
\begin{table*}[!h]
 \caption{Evaluation of OOD detection performance on modified \textbf{20newsgroup dataset} as described in Suppl section A. Note the last column is ECE + Direchlet calibration\cite{kull2019beyond}. Note, our proposed method, PnPOOD demonstrates best OOD detection performance and the least model calibration error. $\pmb{\uparrow}$ indicates larger value is better, and $\pmb{\downarrow}$ indicates lower value is better. All values are percentages. \textbf{Bold} numbers are superior results. }
\label{sample-table}
\vskip 0.15in
\begin{center}
\begin{small}
\begin{sc}
\begin{tabular}{lllccccr}
\toprule
Dataset (IND) & Dataset (OOD) & Method & FPR@90 & AUROC & AUPR & ECE & ECE  \\
\\ 
& & &$\pmb{\downarrow}$ & $\pmb{\uparrow}$ & $\pmb{\uparrow}$ & $\pmb{\downarrow}$  & ($+$ Dir. Cal.) $\pmb{\downarrow}$ 
\\ 

\midrule
\multirow{6}{*}{Computer}
    & \multirow{3}{*}{Sports} & MSP\cite{hendrycks2016baseline} & 0.72 & 0.62 & 0.23 & 0.56 & 0.41 \\ 

& & MSP + ER\cite{hendrycks2018deep} & 0.26 & 0.9 & 0.64 & 0.33 & 0.28 \\

& & MSP + ER + PPLM (PnPOOD) & \textbf{0.18} & \textbf{0.92} & \textbf{0.65} & \textbf{0.32} & \textbf{0.26} \\
  \cmidrule{2-8}  
& \multirow{3}{*}{Politics} & MSP\cite{hendrycks2016baseline} & 0.72 & 0.63 & 0.24 & 0.56 & 0.42 \\ 

& & MSP + ER\cite{hendrycks2018deep} & 0.15 & 0.92 & 0.67 & 0.32 & 0.287 \\

& & MSP + ER + PPLM (PnPOOD) & \textbf{0.11} & \textbf{0.93} & \textbf{0.68} & \textbf{0.31} & \textbf{0.27} \\
\midrule
\multirow{6}{*}{Sports}
    & \multirow{3}{*}{Computer} & MSP\cite{hendrycks2016baseline} & 0.71 & 0.63 & 0.23 & 0.73 & 0.6 \\ 

& & MSP + ER\cite{hendrycks2018deep} & 0.32 & 0.82 & 0.35 & 0.4 & 0.33 \\

& & MSP + ER + PPLM (PnPOOD)& \textbf{0.22} & \textbf{0.89} & \textbf{0.51} & \textbf{0.39} & \textbf{0.31} \\
  \cmidrule{2-8}  
& \multirow{3}{*}{Politics} & MSP\cite{hendrycks2016baseline} & 0.76 & 0.61 & 0.21 & 0.73 & 0.6 \\ 

& & MSP + ER\cite{hendrycks2018deep} & 0.3 & 0.82 & 0.36 & 0.38 & 0.33 \\

& & MSP + ER + PPLM (PnPOOD) & \textbf{0.24} & \textbf{0.87} & \textbf{0.51} & \textbf{0.38} & \textbf{0.32} \\
\midrule
\multirow{6}{*}{Politics}
    & \multirow{3}{*}{Computer} & MSP\cite{hendrycks2016baseline} & 0.61 & 0.67 & 0.25 & 0.72 & 0.61 \\ 

& & MSP + ER\cite{hendrycks2018deep} & 0.24 & 0.91 & 0.64 & 0.48 & 0.41 \\

& & MSP + ER + PPLM (PnPOOD) & \textbf{0.2} &\textbf{0.92} & \textbf{0.6} & \textbf{0.45} & \textbf{0.39} \\
  \cmidrule{2-8}  
& \multirow{3}{*}{Sports} & MSP\cite{hendrycks2016baseline} & 0.63 & 0.67 & 0.25 & 0.71 & 0.62 \\ 

& & MSP + ER\cite{hendrycks2018deep} & 0.42 & 0.85 & 0.53 & 0.47 & 0.41 \\

& & MSP + ER + PPLM (PnPOOD) & \textbf{0.34} & \textbf{0.88} & \textbf{0.56} & \textbf{0.46} & \textbf{0.4} \\
\bottomrule
\end{tabular}
\end{sc}
\end{small}
\end{center}
\end{table*}

Out of the four domains, we choose three \texttt{Computer, Sports and Politics} to train the system leaving out the 'misc' domain due to data leakage issues. We train the system on one domain considering it as ID dataset and test on the remaining domains as OOD datasets. For e.g. We train the system with $D_{ind}$ = {Computer} and $D_{ood}$ as {Sports, Politics} again to overcome issues of data leakage among domains. In this way, we experiment with all the possible  combinations. We evaluate the system on the baseline approaches detailed in Supplemental material:
(a) \textbf{Maximum Softmax Probability (MSP)\cite{hendrycks2016baseline}:} The maximum softmax score is used as a detection score based on the threshold. (b) \textbf{MSP + Entropy Reg. (ER)\cite{hendrycks2018deep}:} A model is trained with the entropy term along with the cross entropy loss as described in Eq. \ref{equ:loss}. The detection score is calculated using the MSP.
(c) \textbf{MSP + ER + PPLM (Our approach-PnPOOD):} A model is trained in a similar manner as in MSP + ER. The samples provided as OOD are generated using the PPLM as described in Section \ref{subsec:dataAugOOD}. The detection score is calculated using the MSP.

 For outlier exposure, we need the train the system with some OOD samples. However, a glaring limitation of the previous state-of-the-art approaches is that they use a generic dataset like Gutenberg\cite{lahiri:2014:SRW} derived from a large collection of books for obtaining OOD samples, and some domains from this dataset may overlap with ID samples resulting in data leakage and misleading performance at test time. To address this the effectiveness of the approach, we use two other domains from the 20Newgroup dataset {Science, Religion} and consider their samples as $D_{ood}^{OE}$ (\cite{hendrycks2018deep}).
 In our approach, instead of using $D_{ood}^{OE}$ drawn from the 20Newsgroup dataset, we generate the samples $D_{ER}^{PPLM}$ using the approach described in  Section ~\ref{sec:methodology}.

 \subsection{Evaluation Metrics}
 \label{subsec:evalMetrics}
 We evaluate our approach for OOD detection employing the standard metrics proposed in OOD literature such as AUROC, AUPR, ECE and FPR@90. We treat the OOD samples as the positive class as proposed by \cite{hendrycks2016baseline}. Refer Appendix \ref{appendix:eval} on details of evaluation metrics.

 \section{Results}
We evaluated our approach on the metrics described in Section ~\ref{sec:experiments}. We present the results on the OOD detection performance when $D_{ind}$ is SST dataset and $D_{ood}$ is MULTI30K, SNLI respectively. Our results demonstrate that our approach outperforms the other state-of-the-art approaches by a significant margin across all the metrics. We want to highlight that previous approaches have exploited the generic dataset like Gutenberg while training with entropy regularization. We believe that $D_{ood}$ should be drawn from a different domain to avoid data leakage. Detailed description and results are present in Appendix \ref{appendix:res} (See Tables on AUROC, AUPR, FPR@90TPR, ECE, ECE+Direchlet Calibration on the benchmark datasets ).

%  \todo{ Lovekesh, Mrinal, In this paper we argue that these prior attempts do not fully address the OOD problem and may suffer from data leakage and poor calibration of the resulting models. Where is the justification of data leakage: Please mention the dataset}

To address the data leakage issues, we experiment with 20Newsgroup dataset. Table \ref{sample-table} illustrates the comparison of our approach with the previous state-of-the-art approaches. Our method outperforms all the other baselines with a significant margins in all the domains. We also report the Expected Calibration Error (ECE) before and after posthoc  Dirichlet calibration.

% Our model was able to give lower score to the OOD sample.

%  \ramya{Mrinal, please fill up the section. The numbers actually thow light about performance of OOD detectors but failure examples may help to comprehend OOD detection in NLP better}

\section{Conclusion}
In the paper, we proposed PnPOOD, a novel technique to detect the out-of-distribution samples using the pseudo OOD samples generated with the Plug and Play language model. We generate  high quality samples which are closer to the IND sample cluster boundary, thus helping in the improving classification performance for OOD detection. In the future, we want to evaluate our approach on CLINC dataset \cite{DBLP:journals/corr/abs-1909-02027, DBLP:journals/corr/abs-2006-10108} and other NLP tasks like token classification in sequence-to-sequence tasks.  

\nocite{langley00}

\bibliography{example_paper}

\begin{thebibliography}{27}
\providecommand{\natexlab}[1]{#1}
\providecommand{\url}[1]{\texttt{#1}}
\expandafter\ifx\csname urlstyle\endcsname\relax
  \providecommand{\doi}[1]{doi: #1}\else
  \providecommand{\doi}{doi: \begingroup \urlstyle{rm}\Url}\fi

\bibitem[Bowman et~al.(2015)Bowman, Angeli, Potts, and Manning]{snli:emnlp2015}
Bowman, S.~R., Angeli, G., Potts, C., and Manning, C.~D.
\newblock A large annotated corpus for learning natural language inference.
\newblock In \emph{Proceedings of the 2015 Conference on Empirical Methods in
  Natural Language Processing (EMNLP)}. Association for Computational
  Linguistics, 2015.

\bibitem[Dathathri et~al.(2020)Dathathri, Madotto, Lan, Hung, Frank, Molino,
  Yosinski, and Liu]{DBLP:journals/corr/abs-1912-02164}
Dathathri, S., Madotto, A., Lan, J., Hung, J., Frank, E., Molino, P., Yosinski,
  J., and Liu, R.
\newblock Plug and play language models: {A} simple approach to controlled text
  generation.
\newblock \emph{ICLR 2020}, abs/1912.02164, 2020.
\newblock URL \url{https://openreview.net/pdf?id=H1edEyBKDS}.

\bibitem[Devlin et~al.(2018)Devlin, Chang, Lee, and
  Toutanova]{DBLP:journals/corr/abs-1810-04805}
Devlin, J., Chang, M., Lee, K., and Toutanova, K.
\newblock {BERT:} pre-training of deep bidirectional transformers for language
  understanding.
\newblock \emph{CoRR}, abs/1810.04805, 2018.
\newblock URL \url{http://arxiv.org/abs/1810.04805}.

\bibitem[Elliott et~al.(2016)Elliott, Frank, Sima'an, and Specia]{W16-3210}
Elliott, D., Frank, S., Sima'an, K., and Specia, L.
\newblock Multi30k: Multilingual english-german image descriptions.
\newblock In \emph{Proceedings of the 5th Workshop on Vision and Language},
  pp.\  70--74. Association for Computational Linguistics, 2016.
\newblock \doi{10.18653/v1/W16-3210}.
\newblock URL \url{http://www.aclweb.org/anthology/W16-3210}.

\bibitem[Gal \& Ghahramani(2016)Gal and Ghahramani]{gal2016dropout}
Gal, Y. and Ghahramani, Z.
\newblock Dropout as a bayesian approximation: Representing model uncertainty
  in deep learning.
\newblock In \emph{international conference on machine learning}, pp.\
  1050--1059. PMLR, 2016.

\bibitem[Guo et~al.(2017)Guo, Pleiss, Sun, and
  Weinberger]{10.5555/3305381.3305518}
Guo, C., Pleiss, G., Sun, Y., and Weinberger, K.~Q.
\newblock On calibration of modern neural networks.
\newblock In \emph{Proceedings of the 34th International Conference on Machine
  Learning - Volume 70}, ICML'17, pp.\  1321–1330. JMLR.org, 2017.

\bibitem[Hein et~al.(2019)Hein, Andriushchenko, and Bitterwolf]{hein2019relu}
Hein, M., Andriushchenko, M., and Bitterwolf, J.
\newblock Why relu networks yield high-confidence predictions far away from the
  training data and how to mitigate the problem.
\newblock In \emph{Proceedings of the IEEE/CVF Conference on Computer Vision
  and Pattern Recognition}, pp.\  41--50, 2019.

\bibitem[Hendrycks \& Dietterich(2019)Hendrycks and
  Dietterich]{hendrycks2018benchmarking}
Hendrycks, D. and Dietterich, T.
\newblock Benchmarking neural network robustness to common corruptions and
  perturbations.
\newblock In \emph{International Conference on Learning Representations}, 2019.
\newblock URL \url{https://openreview.net/forum?id=HJz6tiCqYm}.

\bibitem[Hendrycks \& Gimpel(2017)Hendrycks and Gimpel]{hendrycks2016baseline}
Hendrycks, D. and Gimpel, K.
\newblock A baseline for detecting misclassified and out-of-distribution
  examples in neural networks.
\newblock \emph{International Conference on Learning Representations (ICLR)},
  2017.

\bibitem[Hendrycks et~al.(2019)Hendrycks, Mazeika, and
  Dietterich]{hendrycks2018deep}
Hendrycks, D., Mazeika, M., and Dietterich, T.
\newblock Deep anomaly detection with outlier exposure.
\newblock \emph{International Conference on Learning Representations (ICLR)},
  2019.

\bibitem[Hinton et~al.(2015)Hinton, Vinyals, and Dean]{hinton2015distilling}
Hinton, G., Vinyals, O., and Dean, J.
\newblock Distilling the knowledge in a neural network.
\newblock \emph{arXiv preprint arXiv:1503.02531}, 2015.

\bibitem[Kull et~al.(2019)Kull, Perello-Nieto, K{\"a}ngsepp, Song, Flach,
  et~al.]{kull2019beyond}
Kull, M., Perello-Nieto, M., K{\"a}ngsepp, M., Song, H., Flach, P., et~al.
\newblock Beyond temperature scaling: Obtaining well-calibrated multiclass
  probabilities with dirichlet calibration.
\newblock \emph{arXiv preprint arXiv:1910.12656}, 2019.

\bibitem[Lahiri(2014)]{lahiri:2014:SRW}
Lahiri, S.
\newblock {Complexity of Word Collocation Networks: A Preliminary Structural
  Analysis}.
\newblock In \emph{Proceedings of the Student Research Workshop at the 14th
  Conference of the European Chapter of the Association for Computational
  Linguistics}, pp.\  96--105, Gothenburg, Sweden, April 2014. Association for
  Computational Linguistics.
\newblock URL \url{http://www.aclweb.org/anthology/E14-3011}.

\bibitem[Lakshminarayanan et~al.(2017)Lakshminarayanan, Pritzel, and
  Blundell]{10.5555/3295222.3295387}
Lakshminarayanan, B., Pritzel, A., and Blundell, C.
\newblock Simple and scalable predictive uncertainty estimation using deep
  ensembles.
\newblock In \emph{Proceedings of the 31st International Conference on Neural
  Information Processing Systems}, NIPS'17, pp.\  6405–6416, Red Hook, NY,
  USA, 2017. Curran Associates Inc.
\newblock ISBN 9781510860964.

\bibitem[Lang(1995)]{Lang95}
Lang, K.
\newblock Newsweeder: Learning to filter netnews.
\newblock In \emph{Proceedings of the Twelfth International Conference on
  Machine Learning}, pp.\  331--339, 1995.

\bibitem[Langley(2000)]{langley00}
Langley, P.
\newblock Crafting papers on machine learning.
\newblock In Langley, P. (ed.), \emph{Proceedings of the 17th International
  Conference on Machine Learning (ICML 2000)}, pp.\  1207--1216, Stanford, CA,
  2000. Morgan Kaufmann.

\bibitem[Larson et~al.(2019)Larson, Mahendran, Peper, Clarke, Lee, Hill,
  Kummerfeld, Leach, Laurenzano, Tang, and
  Mars]{DBLP:journals/corr/abs-1909-02027}
Larson, S., Mahendran, A., Peper, J.~J., Clarke, C., Lee, A., Hill, P.,
  Kummerfeld, J.~K., Leach, K., Laurenzano, M.~A., Tang, L., and Mars, J.
\newblock An evaluation dataset for intent classification and out-of-scope
  prediction.
\newblock \emph{CoRR}, abs/1909.02027, 2019.
\newblock URL \url{http://arxiv.org/abs/1909.02027}.

\bibitem[Lee et~al.(2017)Lee, Lee, Lee, and Shin]{lee2017training}
Lee, K., Lee, H., Lee, K., and Shin, J.
\newblock Training confidence-calibrated classifiers for detecting
  out-of-distribution samples.
\newblock \emph{arXiv preprint arXiv:1711.09325}, 2017.

\bibitem[Lee et~al.(2018)Lee, Lee, Lee, and Shin]{10.5555/3327757.3327819}
Lee, K., Lee, K., Lee, H., and Shin, J.
\newblock A simple unified framework for detecting out-of-distribution samples
  and adversarial attacks.
\newblock In \emph{Proceedings of the 32nd International Conference on Neural
  Information Processing Systems}, NIPS'18, pp.\  7167–7177, Red Hook, NY,
  USA, 2018. Curran Associates Inc.

\bibitem[Liang et~al.(2017)Liang, Li, and Srikant]{liang2017enhancing}
Liang, S., Li, Y., and Srikant, R.
\newblock Enhancing the reliability of out-of-distribution image detection in
  neural networks.
\newblock \emph{arXiv preprint arXiv:1706.02690}, 2017.

\bibitem[Liu et~al.(2020)Liu, Lin, Padhy, Tran, Bedrax{-}Weiss, and
  Lakshminarayanan]{DBLP:journals/corr/abs-2006-10108}
Liu, J.~Z., Lin, Z., Padhy, S., Tran, D., Bedrax{-}Weiss, T., and
  Lakshminarayanan, B.
\newblock Simple and principled uncertainty estimation with deterministic deep
  learning via distance awareness.
\newblock \emph{CoRR}, abs/2006.10108, 2020.
\newblock URL \url{https://arxiv.org/abs/2006.10108}.

\bibitem[Patel et~al.(2021)Patel, Beluch, Zhang, Pfeiffer, and
  Yang]{patel2021manifold}
Patel, K., Beluch, W., Zhang, D., Pfeiffer, M., and Yang, B.
\newblock On-manifold adversarial data augmentation improves uncertainty
  calibration.
\newblock In \emph{2020 25th International Conference on Pattern Recognition
  (ICPR)}, pp.\  8029--8036. IEEE, 2021.

\bibitem[Radford et~al.(2019)Radford, Wu, Child, Luan, Amodei, and
  Sutskever]{radford2019language}
Radford, A., Wu, J., Child, R., Luan, D., Amodei, D., and Sutskever, I.
\newblock Language models are unsupervised multitask learners.
\newblock 2019.

\bibitem[Ren et~al.(2019)Ren, Liu, Fertig, Snoek, Poplin, Depristo, Dillon, and
  Lakshminarayanan]{NEURIPS2019_1e795968}
Ren, J., Liu, P.~J., Fertig, E., Snoek, J., Poplin, R., Depristo, M., Dillon,
  J., and Lakshminarayanan, B.
\newblock Likelihood ratios for out-of-distribution detection.
\newblock In Wallach, H., Larochelle, H., Beygelzimer, A., d\' Alch\'{e}-Buc,
  F., Fox, E., and Garnett, R. (eds.), \emph{Advances in Neural Information
  Processing Systems}, volume~32. Curran Associates, Inc., 2019.
\newblock URL
  \url{https://proceedings.neurips.cc/paper/2019/file/1e79596878b2320cac26dd792a6c51c9-Paper.pdf}.

\bibitem[Socher et~al.(2013)Socher, Perelygin, Wu, Chuang, Manning, Ng, and
  Potts]{socher-etal-2013-recursive}
Socher, R., Perelygin, A., Wu, J., Chuang, J., Manning, C.~D., Ng, A., and
  Potts, C.
\newblock Recursive deep models for semantic compositionality over a sentiment
  treebank.
\newblock In \emph{Proceedings of the 2013 Conference on Empirical Methods in
  Natural Language Processing}, pp.\  1631--1642, Seattle, Washington, USA,
  October 2013. Association for Computational Linguistics.
\newblock URL \url{https://www.aclweb.org/anthology/D13-1170}.

\bibitem[Wolf et~al.(2020)Wolf, Debut, Sanh, Chaumond, Delangue, Moi, Cistac,
  Rault, Louf, Funtowicz, Davison, Shleifer, von Platen, Ma, Jernite, Plu, Xu,
  Scao, Gugger, Drame, Lhoest, and Rush]{wolf-etal-2020-transformers}
Wolf, T., Debut, L., Sanh, V., Chaumond, J., Delangue, C., Moi, A., Cistac, P.,
  Rault, T., Louf, R., Funtowicz, M., Davison, J., Shleifer, S., von Platen,
  P., Ma, C., Jernite, Y., Plu, J., Xu, C., Scao, T.~L., Gugger, S., Drame, M.,
  Lhoest, Q., and Rush, A.~M.
\newblock Transformers: State-of-the-art natural language processing.
\newblock In \emph{Proceedings of the 2020 Conference on Empirical Methods in
  Natural Language Processing: System Demonstrations}, Online, October 2020.
  Association for Computational Linguistics.

\bibitem[Zhang et~al.(2006)Zhang, Berg, Maire, and Malik]{zhang2006svm}
Zhang, H., Berg, A.~C., Maire, M., and Malik, J.
\newblock Svm-knn: Discriminative nearest neighbor classification for visual
  category recognition.
\newblock In \emph{2006 IEEE Computer Society Conference on Computer Vision and
  Pattern Recognition (CVPR'06)}, volume~2, pp.\  2126--2136. IEEE, 2006.

\end{thebibliography}
\bibliographystyle{icml2021}
\clearpage
\newpage

\icmltitlerunning{Supplementary: ICML 2021 Workshop on Uncertainty and Robustness in Deep Learning}
\appendix

\section{Evaluation metrics}
\label{appendix:eval}

The following are description on the OOD detection metrics:

\begin{itemize}
    \item \textbf{AUROC:} The area under the receiver operating characteristic (AUROC) is a common metric used in OOD detection,  specifically, it is the area under the FPR against TPR curve.  Higher value represents a better model.
    \item \textbf{AUPR:} The area under the precision-recall curve is used. Higher value represents better model. Particularly useful incase of ID-OOD sample imbalance.
    \item \textbf{FPR@90:}  False positive rate (FPR) at $90\%$ true positive rate. Lower is better.
    \item \textbf{ECE:} The Expected Calibration Error (ECE) takes a weighted average over the difference of absolute accuracy and confidence/prediction probability. Lower is preferred~\cite{10.5555/3305381.3305518}.
    \begin{equation}
        ECE = \sum\limits_{b=1}^B \frac{n_b}{N}  |acc(b) - conf(b)|,
    \end{equation}
    
    where $n_{b}$ is the number of predictions in bin $b$, $N$ is the total number of data points, and acc(b) and conf(b) are the accuracy and confidence of bin $b$, respectively.
     
\end{itemize}

\section{Dataset description of 20newsgroup}
\label{sec:20newsgrp}

Table \ref{stats} illustrates the detailed information about the domains and classes present in 20Newsgroup dataset. We want to highlight that we leave out Domain : 4 (misc) as it includes samples from other domains resulting in data leakage. 

\begin{table}[ht]
\caption{Dataset Description of 20newsgroup}
\label{stats}
\vskip 0.15in
\begin{center}

\begin{small}
\begin{sc}
     \scalebox{0.65}{
\begin{tabular}{l||c||r}
\toprule
\midrule
\multirow{6}{*}{}
    & \multirow{6}{*}{} &  \multirow{6}{*}{}\\ 
    \textbf{Domain 1: Computer} & \textbf{Domain 2: Sports} & \textbf{Domain 3: Science}\\
    comp.graphics &  & \\
comp.os.ms-windows.misc  & rec.autos &sci.crypt\\
comp.sys.ibm.pc.hardware & rec.motorcycles &sci.electronics\\
comp.sys.mac.hardware & rec.sport.baseball &sci.med\\
comp.windows.x	 & rec.sport.hockey	 &sci.space\\

\midrule
\midrule

\multirow{4}{*}{}
    & \multirow{4}{*}{} &  \multirow{4}{*}{}\\ 
        \textbf{Domain 4: Misc.} & \textbf{Domain 5: Politics} & \textbf{Domain 6: Religion}\\
 & talk.politics.misc &talk.religion.misc\\
misc.forsale & talk.politics.guns &alt.atheism\\
 & talk.politics.mideast  &soc.religion.christian\\

\bottomrule
\bottomrule
\end{tabular}}
\end{sc}
\end{small}
\end{center}
\vskip -0.1in
\end{table}

\begin{table*}[ht]
\caption{Evaluation of OOD detection performance on \textbf{SST dataset} as in-distribution. Note, our proposed method, PnPOOD demonstrates best OOD detection performance and the least model calibration error. $\pmb{\uparrow}$ indicates larger value is better, and $\pmb{\downarrow}$ indicates lower value is better. All values are percentages. \textbf{Bold} numbers are superior results.}
\label{results:sst}
\vskip 0.15in
\begin{center}
\begin{small}
\begin{sc}
\begin{tabular}{lllccr}
\toprule
Dataset (IND) & Dataset (OOD) & Method & FPR@90 & AUROC & AUPR  \\
\\ 
& & &$\pmb{\downarrow}$ & $\pmb{\uparrow}$ & $\pmb{\uparrow}$  
\\ 

\midrule
\multirow{6}{*}{SST}
    & \multirow{3}{*}{Multi30K} & MSP\cite{hendrycks2016baseline} & 0.85 & 0.54 & 0.19  \\ 

& & MSP + ER\cite{hendrycks2018deep}& 0.47 & 0.81 & 0.44 \\

& & MSP + ER + PPLM (PnPOOD) & \textbf{0.40} & \textbf{0.84} & \textbf{0.48}  \\
  \cmidrule{2-6}  
& \multirow{3}{*}{SNLI} & MSP\cite{hendrycks2016baseline} & 0.63 & 0.73 & 0.32  \\ 

& & MSP + ER\cite{hendrycks2018deep} & 0.11 & 0.95 & 0.71  \\

& & MSP + ER + PPLM (PnPOOD) & \textbf{0.05} & \textbf{0.97} & \textbf{0.77}  \\
\bottomrule
\end{tabular}
\end{sc}
\end{small}
\end{center}
\vskip -0.1in
\end{table*}

\section{Setup and Results}
\label{appendix:res}

\subsection{Configuration Details}
 We conduct experiment with a LSTM based text classifier. We stacked two LSTM layers of dimension $128$. We initialize the input embeddings with pre-trained GloVe\footnote{https://nlp.stanford.edu/projects/glove/} vectors of size $300$. We used the Adam optimizer with learning rate as $0.001$ and drop out set to $0.3$. The batch size is $32$. We perform our experiments on NVidia GTX 1070 with 16 GB RAM. All the models are implemented in Pytorch.
 
 \subsection{Results}
 
 Table \ref{results:sst} demonstrates the results on the SST dataset. Our method outperforms previous state-of-the-art approaches by a significant margin.
 
  \begin{figure*}
\centering
\subfigure[MSP]{
\includegraphics[width=.3\textwidth]{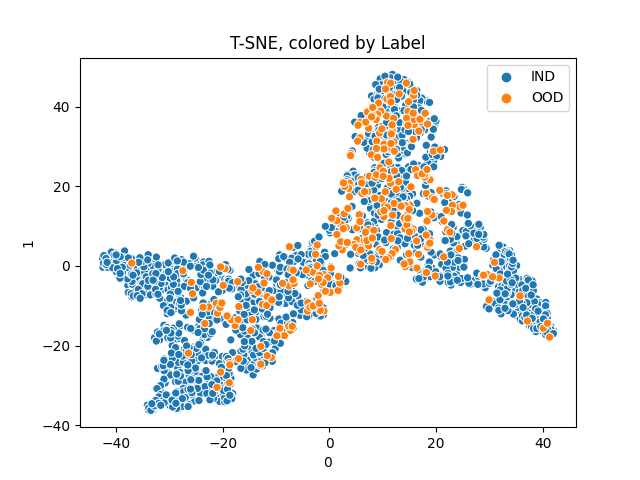}
}
\subfigure[MSP + ER]{
\includegraphics[width=.3\textwidth]{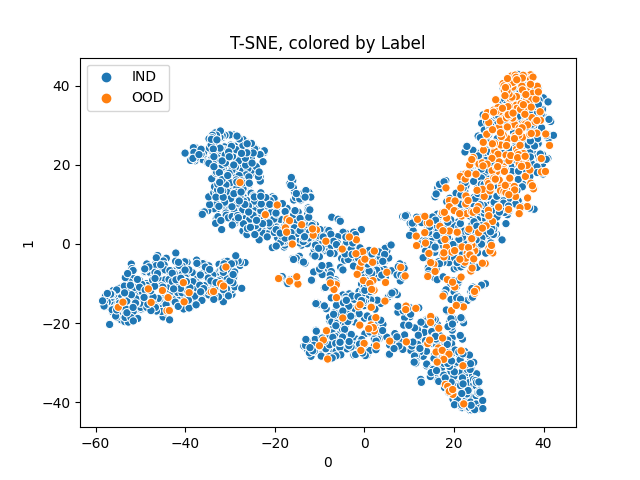}
}
\subfigure[MSP + ER + PPLM (PnPOOD)]{
\includegraphics[width=.3\textwidth]{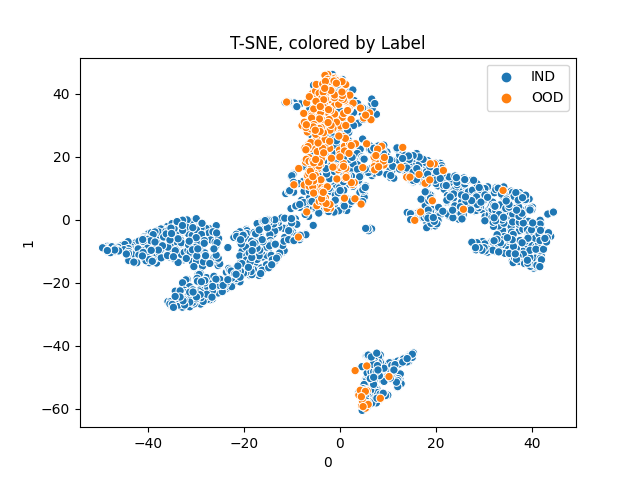}
}
\caption{t-SNE plot illustrating that our proposed approach yields better separation between IND and OOD samples on a pair of domains in 20newsgroup dataset.}
\label{fig:tsne}
\end{figure*}

\begin{figure*}[!h]
\centering
\subfigure[$D_{ind}$ = Sports \& $D_{ood}$ = Computer]{
\includegraphics[width=.45\textwidth]{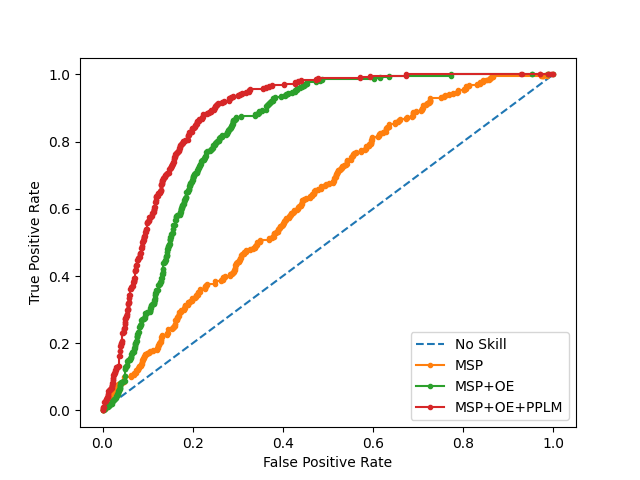}
}
\subfigure[$D_{ind}$ = Sports \& $D_{ood}$ = Politics]{
\includegraphics[width=.45\textwidth]{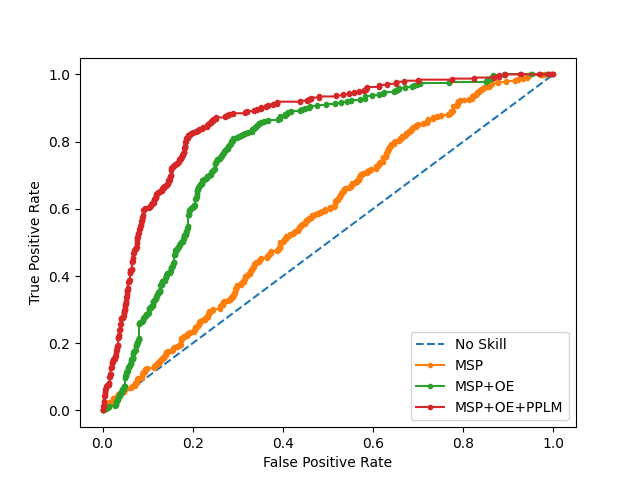}
}
\caption{AUROC on different pairs of domains of 20Newsgroup, (a) AUROC when $D_{ind} = Sports$ and $D_{ood}$ = Computer (b) AUROC when $D_{ind} = Sports$ and $D_{ood}$ = Politics. Note:  AUROC is a plot of False Positive Rate (x-axis) and True Positive Rate (y-axis), it shows the performance of classification models under all thresholds.}
\label{fig:auroc}
\end{figure*}

Figure \ref{fig:auroc} shows the AUROC plot of sports domain as $D_{ind}$ when $D_{ood}$ is computer, politics respectively. Observe our proposed approach yields a higher AUROC in comparison to competing methods after Direchlet Calibration.

Figure \ref{fig:tsne} illustrates the t-SNE plot of our method w.r.t. to other baselines. We demonstrate that the samples generated using our approach yields a better OOD classification performance visually as depicted in t-SNE plot in Suppl. material.

Table \ref{results:in_acc} illustrates the in-distribution accuracy on 20Newsgroups dataset. We report results in comparison with previous baselines. We observe that including the entropy regularization did not affect the in-distribution classifier's performance.

\begin{table}[h]
\caption{Evaluation of IND classifier accuracy on \textbf{20Newsgroups dataset}. $\pmb{\uparrow}$ indicates larger value is better, and $\pmb{\downarrow}$ indicates lower value is better. All values are percentages. \textbf{Bold} numbers are superior results.}
\label{results:in_acc}
\vskip 0.15in
\begin{center}
\begin{small}
\begin{sc}
\begin{tabular}{lcr}
\toprule
Dataset & Method & Acc  \\
\\ 
& &$\pmb{\uparrow}$  
\\ 

\midrule
\multirow{3}{*}{Computer} & MSP\cite{hendrycks2016baseline} & 0.46  \\

& MSP + ER\cite{hendrycks2018deep}& 0.47 \\

& MSP + ER + PPLM (PnPOOD) & \textbf{0.48} \\
  \cmidrule{1-3}  
 \multirow{3}{*}{Sports} & MSP\cite{hendrycks2016baseline} & 0.73  \\ 

& MSP + ER\cite{hendrycks2018deep} & 0.73  \\

& MSP + ER + PPLM (PnPOOD) & \textbf{0.74}   \\
  \cmidrule{1-3}  
 \multirow{3}{*}{Politics} & MSP\cite{hendrycks2016baseline} & \textbf{0.65}  \\ 

& MSP + ER\cite{hendrycks2018deep} & 0.64  \\

& MSP + ER + PPLM (PnPOOD) & 0.64   \\
\bottomrule
\end{tabular}
\end{sc}
\end{small}
\end{center}
\vskip -0.1in
\end{table}

 \section{Example success and failure cases}
\label{appendix:snf}
We train the classifier with  $D_{ind}$ as Computer and tested on the samples from $D_{ood}$ as Sports. The following examples shows the success and failure cases:

\textbf{Success Case}:

\textbf{Test data:} ''Headlights problem thanks all you who responded posting the problem with truck headlights low beam problem was loose wire connection was not the fuse minority you suggested thanks again''

\begin{table}[!h]
 \caption{Softmax scores}
\label{success_score}
\vskip 0.15in
\begin{center}
\begin{small}
\begin{sc}
\begin{tabular}{l|r}
\textbf{Method} & \textbf{Softmax score} \\
& $\pmb{\downarrow}$\\
\toprule
MSP\cite{hendrycks2016baseline}r & 0.59 \\
MSP + ER\cite{hendrycks2018deep} & 0.26 \\
MSP + ER + PPLM (PnPOOD) & \textbf{0.19} \\
\bottomrule
\end{tabular}
\end{sc}
\end{small}
\end{center}
\vskip -0.1in
\end{table}

Our model was able to give lower score to the OOD test sample input to the classifier. Refer Table \ref{success_score}.

\textbf{Failure Case}:

\textbf{Text:} ``how hard change springs truck article apr michael apple com ems michael apple com michael smith writes does take any peculiar tools remove the rear springs from ford truck naah just coupla nice big bumps''

\begin{table}[!h]
 \caption{Softmax scores}
\label{failure_score}
\vskip 0.15in
\begin{center}
\begin{small}
\begin{sc}
\begin{tabular}{l|r}
\textbf{Method} & \textbf{Softmax score} \\
& $\pmb{\downarrow}$\\
\toprule
MSP\cite{hendrycks2016baseline} & \textbf{0.65} \\
MSP + ER\cite{hendrycks2018deep} & 0.84 \\
MSP + ER + PPLM (PnPOOD) & 0.98 \\
\bottomrule
\end{tabular}
\end{sc}
\end{small}
\end{center}
\vskip -0.1in
\end{table}

On closer inspection, we believe that the words like ``apple'', ``.com'' are from computer domain and may have confused the classifier. Refer Table \ref{failure_score}

\section{Algorithm for OOD sample generation}
\label{appendix:algo}

 We generate OOD samples using PPLM~\cite{DBLP:journals/corr/abs-1912-02164} that utilize an attribute model in order to guide generation without fine tuning the large Natural language generation model. We also perform filtering of samples which are closer in the embedding space to the IND cluster boundary are more relevant for discriminating between IND and OOD samples. We conducted multiple experiments setting the threshold in the [0,5,10,12,20,25] and setting a threshold value as 10, we obtain the maximum OOD performance. We observed that increasing the threshold value greater than 10 increased the number of generated sentences and we observed the OOD performance reduced with increase in threshold. Similarly, by decreasing the threshold we were losing out on important OOD sentences. 

\begin{algorithm}[!h]
	\label{algo_PeaPOOD}
	\SetKwInput{KwInput}{Inputs}                % Set the Input
	\SetKwInput{KwOutput}{Output}  
	\SetKwInput{KwLoss}{Loss}  
	\DontPrintSemicolon
	\LinesNotNumbered
	\KwInput{
	
		$\D_{ood}^{PPLM}$    \tcp*{OOD Samples}
		$\D_{ind}$    \tcp*{IND Samples}
		$T$ = 10    \tcp*{Threshold Distance}
	}

	\SetKwFunction{FSum}{ClusterDistance}
	
	\SetKwProg{Fn}{Function}{}{}
	\Fn{\FSum{E}}{
		
		$C \gets \frac{1}{N} \sum\limits_{i=1}^N E_i$ \tcp*{finds cluster center of embeddings}
		
		$Dist \gets \emptyset$ \\
		\ForEach{$e$ in $E$}{
		    $Dist_j \gets $ $EucDistance$({$C$,  $e$}) \tcp*{Euclidean distance}
		}
		
		$d \gets  percentile(Dist, 0.95)$   \tcp{remove outliers}
    % 		$d \gets \sum\limits_{i=1}^N |C - E_i|$ \tcp*{finds cluster center of embeddings}

		\KwRet C, d 
	}
	\tcp{Training}
         $E_{ood} \gets $ BertEmbeddings($D_{ood}^{PPLM})$ \\
		$E_{ind} \gets BertEmbeddings(D_{ind})$ \\
	    $C, d \gets $ \FSum{$E_{ind}$} \tcp*{Distance from cluster center to the boundary}
	
	    $S \gets \emptyset$  \tcp*{Filtered Sentences}
	    
		\ForEach{ $s_i$  in $\D^\text{PPLM}_\text{ood}$}{          
			\If{  $d$  < $EucDistance$({$C$, $s_{i}$}) < $d$ + $T$ }{
                $S_j \gets s_i$;
            }
		}
		
		\KwRet $S$ 
	\caption{Filtering the $D_{ood}^{PPLM}$ samples} 
	\label{algo_PeaPOOD}
\end{algorithm}

%%%%%%%%%%%%%%%%%%%%%%%%%%%%%%%%%%%%%%%%%%%%%%%%%%%%%%%%%%%%%%%%%%%%%%%%%%%%%%%
%%%%%%%%%%%%%%%%%%%%%%%%%%%%%%%%%%%%%%%%%%%%%%%%%%%%%%%%%%%%%%%%%%%%%%%%%%%%%%%
% DELETE THIS PART. DO NOT PLACE CONTENT AFTER THE REFERENCES!
%%%%%%%%%%%%%%%%%%%%%%%%%%%%%%%%%%%%%%%%%%%%%%%%%%%%%%%%%%%%%%%%%%%%%%%%%%%%%%%
%%%%%%%%%%%%%%%%%%%%%%%%%%%%%%%%%%%%%%%%%%%%%%%%%%%%%%%%%%%%%%%%%%%%%%%%%%%%%%%
%\appendix
%\section{Do \emph{not} have an appendix here}
%
%\textbf{\emph{Do not put content after the references.}}
%%
%Put anything that you might normally include after the references in a separate
%supplementary file.
%
%We recommend that you build supplementary material in a separate document.
%If you must create one PDF and cut it up, please be careful to use a tool that
%doesn't alter the margins, and that doesn't aggressively rewrite the PDF file.
%pdftk usually works fine. 
%
%\textbf{Please do not use Apple's preview to cut off supplementary material.} In
%previous years it has altered margins, and created headaches at the camera-ready
%stage. 
%%%%%%%%%%%%%%%%%%%%%%%%%%%%%%%%%%%%%%%%%%%%%%%%%%%%%%%%%%%%%%%%%%%%%%%%%%%%%%%
%%%%%%%%%%%%%%%%%%%%%%%%%%%%%%%%%%%%%%%%%%%%%%%%%%%%%%%%%%%%%%%%%%%%%%%%%%%%%%%

\end{document}